\documentclass[a4paper]{article}
\usepackage{graphicx}
\usepackage{onecolceurws}
\usepackage{multirow}

\title{Batch Clustering for Multilingual News Streaming}

\author{
Mathis Linger \\ linger.mathis@gmail.com
\and
Mhamed Hajaiej \\ mhamed.hajaiej@gmail.com
}

\institution{BNP Paribas AI, Paris, France}

\begin{document}
\maketitle

\begin{abstract}
Nowadays, digital news articles are widely available, published by various editors and often written in different languages. This large volume of diverse and unorganized information makes human reading very difficult or almost impossible. This leads to a need for algorithms able to arrange high amount of multilingual news into stories. To this purpose, we extend previous works on Topic Detection and Tracking, and propose a new system inspired from \textit{newsLens}. We process articles per batch, looking for monolingual local topics which are then linked across time and languages. Here, we introduce a novel "replaying" strategy to link monolingual local topics into stories. Besides, we propose new fine tuned multilingual embedding using SBERT to create crosslingual stories. Our system gives monolingual state-of-the-art results on dataset of Spanish and German news and crosslingual state-of-the-art results on English, Spanish and German news.
\end{abstract}
\vskip 32pt

\section{Introduction}
The rise of the Internet and social media has increased the number of news articles available. This flow of information is difficult to ingest but can yet be very valuable for real-time economics decisions. For instance, within the bank industry, being able to extract the main information from numerous and diverse news articles can help to assess risk linked to companies. This leads to a need for scalable systems, able to organize a large, multi-source and multilingual flow of news articles. The Topic Detection and Tracking (TDT) task refers to techniques that automatically search, organize and structure news from a variety of broadcast news media. More precisely, its aim is to arrange an unordered multilingual stream of news articles into major events clusters called stories.

Recently, a two-steps streaming system called \textit{newsLens} \cite{Laban2017newsLensBA} has been proposed to group articles into stories. However, this system does not support multilingual articles. Other works such as the one introduced by Staykovski \textit{et al.} \cite{Staykovski2019DenseVS}, investigate methods to represent news articles and link them along time. Based on these works, we propose an extention of the \textit{newsLens} algorithm. Experiments on standard benchmark dataset introduced by Miranda \textit{et al.} \cite{Miranda2018MultilingualCO} for news stream clustering show significant improvement over the state-of-the-art for the two main tasks: monolingual and multilingual news clustering.
%Recently, a two-steps streaming system called \textit{newsLens} \cite{Laban2017newsLensBA} has been proposed to arrange articles into stories. However, this system does not support multilingual articles, and the topic matching procedure used to link topics along time can be further improved. To tackle these issues, we propose an extention of the \textit{newsLens} algorithm. Experiments on standard benchmark dataset introduced by Miranda \textit{et al.} \cite{Miranda2018MultilingualCO} for news stream clustering show significant improvement over the state-of-the-art for the two main tasks: monolingual and multilingual news clustering.

Our proposed approach improves topic tracking and handles multilingual data. Specifically, our contribution is \textbf{twofold}: %twofold:
\begin{itemize}
    \item[$\bullet$] We extend \textit{newsLens} \cite{Laban2017newsLensBA} with a new topic matching procedure refered to as "replaying" strategy to link monolingual topics through time.
%    \item[$\bullet$] We show that multilingual documents are best represented through new XLM-RoBERTa embedding model \cite{ROBERTA}.
    \item[$\bullet$] We show that multilingual documents are best represented through fine tuned DistilBERT \cite{Sanh2019DistilBERTAD} multilingual model using SBERT \cite{Reimers2019SentenceBERTSE} triplet network structure.
    %\item[$\bullet$] Solving an optimal assignment problem based on these new representations yields better story clustering across languages.
    %\item[$\bullet$] We link stories across languages using these representations and solving an optimal assignment problem.
\end{itemize}
\section{Related Work}
This paper directly follows a growing body of work on Topic Detection and Tracking (TDT). Most of these works aim to solve the TDT task processing articles according to well defined pipelines.

In a first approach, Laban and Hearst \cite{Laban2017newsLensBA} propose \textit{newsLens}, a two-steps streaming system which first extracts keywords to create topics using Louvain community detection algorithm \cite{Blondel2008FastUO}, and then solidify these local topics clusters into stories by comparing their keywords distribution. Emphasis is put on scalability, as the volume of news articles processed in their proposed experimental setup is about 4 millions. However, the algorithm does not handle multilingual articles and uses simple TF-IDF based method to compare articles. Moreover, its performance was not formally evaluated at the time.

Later, Miranda \textit{et al.} \cite{Miranda2018MultilingualCO} introduce a novel method to cluster an incoming stream of multilingual documents into monolingual and crosslingual stories. Documents are embedded in two latent spaces, a "monolingual space" and a "multilingual space", which are used to cluster articles into topics. Then, new incoming articles can contribute to stories if they are close enough to topics centroids.
Alongside their proposed system, they introduce a multilingual dataset adapted from Rupnik \textit{et al.} \cite{Rupnik2015NewsAL} containing articles in English, Spanish and German which have been manually annotated with monolingual and crosslingual story cluster labels. To the best of our knowlege, no other multilingual benchmark dataset has been proposed for the TDT task.

Recently, Staykovski \textit{et al.} \cite{Staykovski2019DenseVS} use the English part of the corpus from Miranda \textit{et al.} \cite{Miranda2018MultilingualCO} to assess the importance of article representations for news clustering. Among other, they show that sparse vector representation with TF-IDF weighting yields better results than doc2vec-based dense representation \cite{Le2014DistributedRO}.

\section{Our system}

\begin{figure}[ht]
\begin{center}
\includegraphics[height=6.5cm]{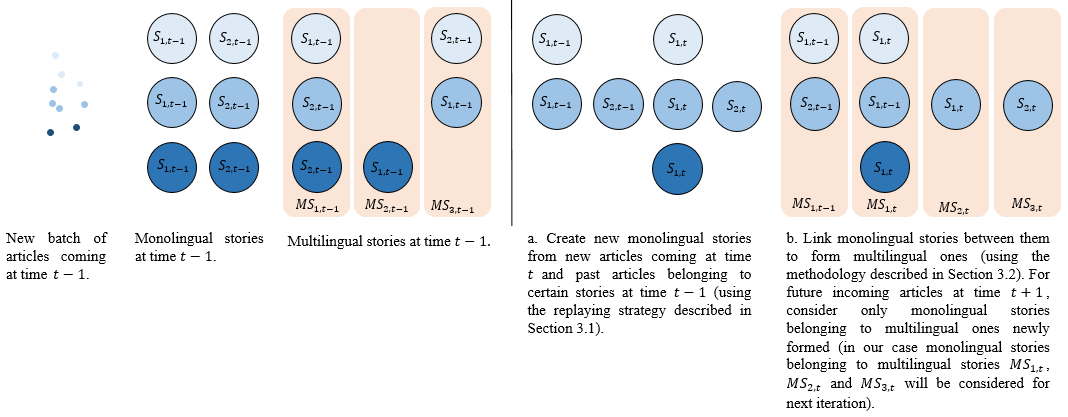}
\caption{General description of our system. Colors represent languages of articles/stories. $S_{i,t}$ represents story $i$ created at time $t$. $MS_{j,t}$ represents multilingual story $j$ created at time $t$.}
\label{fig:figureglobale}
\end{center}
\end{figure}

We extend \textit{newsLens} with a per-batch procedure, where documents published within a close range of time are processed to form local monolingual topics. Monolingual stories are then created by linking topics across time (i.e. across batches, Figure \ref{fig:figureglobale}a) using topic centroids method from Miranda \textit{et al.} \cite{Miranda2018MultilingualCO}. Finally, multilingual stories are created by aggregating monolingual stories from different languages whose representations in a multilingual latent space are close enough (Figure \ref{fig:figureglobale}b).

\bigskip

\subsection{Monolingual stories}\label{monolingualStories}
In order to create local topics, we process articles per batch of close range, computing similarities between each pair of articles and making use of a community detection algorithm. Then, we link the local topics along time thanks to a "replaying" strategy based on topics centroids similarities.

\subsubsection{Article representation}
Following Miranda \textit{et al.} \cite{Miranda2018MultilingualCO} and Staykovski \textit{et al.} \cite{Staykovski2019DenseVS} who demonstrated the inefficiency of dense features to cluster documents of a same language, articles are represented using sparse TF-IDF features. Monolingual representations for each document consist of 9 TF-IDF weighted bag of words sub-vectors, corresponding to the entities, lemmas and tokens contained in the title, body and title+body of each document. Contrary to previous work, we do not use any time feature, time being implicitely taken into account by the per-batch procedure. Please note that in all proposed experiments, we use the same entities, lemmas and tokens as already extracted by Miranda \textit{et al.} \cite{Miranda2018MultilingualCO} to ensure fair comparison of our proposed system. 

\subsubsection{Topics detection}
To group articles into local topics, we build a non-oriented graph, where nodes represent articles and edges are weighted by the similarity between articles. More precisely, weights associated to edges are a linear combination of the cosine similarities between each of the 9 TF-IDF based representative sub-vectors of articles. Formally, the similarity function between two articles $i$ and $j$ is computed as :

\begin{equation}
\label{eq:equation1}
\sum^{K}_{k=0}\beta_k\times\theta(d_i^k,d_j^k)
\end{equation}
Where $K$ is the number of sub-vectors used to represent an article (i.e. $K=9$), $\beta_k$ are learned weights associated to sub-vector $k$, $\theta$ is the cosine similarity function and $d_i^k$ is the sub-vector $k$ of article $i$.

In order to learn the best $\beta$ weights  to aggregate the cosine similarities of articles’ representations for each language, we fit a logistic regression using the training part of the dataset. More precisely, for articles of a same language, we compute the cosine similarities between sub-vectors of each pair of articles. We then assign positive labels for pairs which are indeed of the same story and a negative ones for pairs of different stories.

Besides, using (\ref{eq:equation1}), we are able to compute similarities between each pair of articles. The resulting adjacency matrix can be viewed as a graph displaying weak links for pairs of articles which are of different stories and strong ones for pairs of a same story. We apply the Louvain community detection algorithm \cite{Blondel2008FastUO} to this graph in order to extract well delimited communities of articles, which will be refered to as topics.

\subsubsection{Linking topics through time}

\begin{figure}[ht]
\begin{center}
\includegraphics[height=5.7cm]{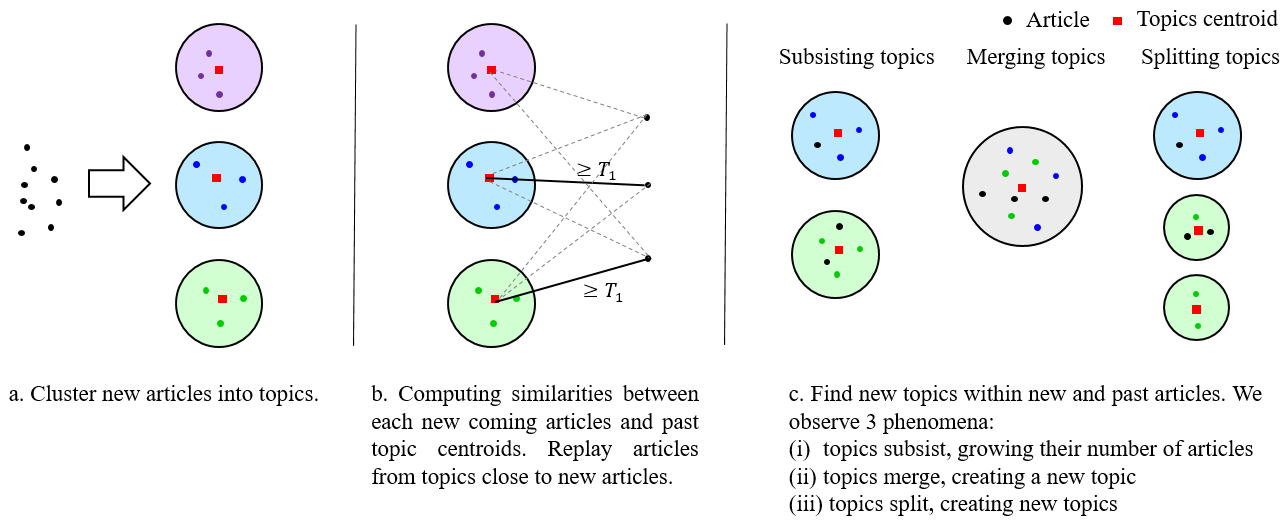}
\caption{Linking topics along time using the "replaying" strategy.}
\label{fig:figure1}
\end{center}
\end{figure}

Once a batch of documents has been clustered into topics, topics are linked across time (i.e. across batches) to form monolingual stories. To this end, we introduce a "replaying" strategy (Figure \ref{fig:figure1}), based on the similarity between articles in the current batch and topic centroids from previous batches. More precisely, when topics are created within a batch of articles at time $t-1$, we compute their centers as the average of each articles representative sub-vectors (Figure \ref{fig:figure1}a). Then, for a new batch of articles at $t$, we compute similarities between all new articles and all topics centers at $t-1$ using the formula introduced in (\ref{eq:equation1}). When a topic at $t-1$ has a similarity with a new article at $t$ greater than a threshold $T_1$ (in our implementation $T_1=0.43$ for English, $0.61$ for German and $0.52$ for Spanish documents), we replay all the articles constituting the topic at $t-1$, i.e. we add those articles to the current batch so that they can be considered during the new round of topic detection at time $t$ (Figure \ref{fig:figure1}b). This approach allows for the emmergence of different topic behaviors across time (Figure \ref{fig:figure1}c): indeed, a previously created topic can subsist (eventually aggregating new articles) or not; it can also be split into several new topics. Finally, two or more topics can be merged into one.

%\begin{figure}[ht]
%\includegraphics[width=\textwidth]{Figure_1bis.png}
%\caption{Linking topics along time using the "replaying" strategy.} %\label{fig:figure1}
%\end{figure}

\subsection{From monolingual to multilingual stories}\label{fromMonoToMulti}
Whenever we create new monolingual stories, we try to link them with current and past stories in other languages. To do so, we compute a common representation for stories in different languages, and associate them solving an optimal assignment problem.

\subsubsection{Story representation}
%We represent each story by one vector using pre-trained XLM-RoBERTa base multilingual embeddings \cite{ROBERTA}. More precisely, we first compute the embedding for each article by processing the concatenation of their title and body.
To represent each article in a multilingual space, we use the SBERT \cite{Reimers2019SentenceBERTSE} triplet network structure. We use the training part of our dataset in order to create labeled sentence triplets: The anchor and the positive example are articles in different languages coming from the same story, while the negative example is an article in different language and story than the anchor. We fine tune the multilingual DistilBERT \cite{Sanh2019DistilBERTAD} model using the concatenation of title and body articles for 15 epochs on 6,000 semi-hard triplets selected at the beginning of each epoch. We use a batch size of 8, a gradient accumulation of 2 steps, Adam optimizer with learning rate $2e-5$ and using the MEAN pooling strategy. Then, in order to get the representation of a monolingual story, we average these representations over all articles within the story.

\begin{equation}
\label{eq:equation2}
s_i = \frac{1}{|S_i|}\sum_{j \in S_i}e_j
\end{equation}

Where $S_i$ is story $i$, and $e_j$ is our fine tuned multilingual DistilBERT embedding corresponding to article $j$.

\subsubsection{Linking monolingual into multilingual stories}
Based on the conclusions of Miranda \textit{et al.} \cite{Miranda2018MultilingualCO}, we use English as a pivot language in order to link monolingual into crosslingual stories. More precisely, we compute cosine similarities between non English and English story embeddings.
Assuming that, at maximum one story of a given language can contribute to a multilingual story, we have to solve an optimal assignment problem. Between two sets of stories in different languages, we have to find the stories assignment between the two languages such that the sum of the similarities (resp. distances) of the linked story pairs is the highest (resp. the lowest) possible. We use the Hungarian algorithm \cite{Kuhn1955Hungarian} to solve this problem in polynomial time. More precisely, we define the cost function as the distance matrix (1 - similarity matrix) between stories of two different languages. Since some stories may not be related between two languages, we allow assignments only if the distance between two stories is less than a threshold $T_2$ ( $T_2$ is set to 0.22 in our implementation). We make this connection between monolingual stories each time we receive a new batch of articles, taking into account all monolingual stories not already assigned to multilingual stories and not older than 4 batches.

%\begin{figure}
%\includegraphics[width=\textwidth]{Figure_2.PNG}
%\caption{Example of optimal assignement problem. Stories A and A' as well as B and B' will form multilingual stories.} \label{fig:figure2}
%\end{figure}

\section{Experimental setup}

\subsection{Dataset}

We assess the effectiveness of our proposed approach on the standard multilingual dataset introduced by Miranda \textit{et al.} \cite{Miranda2018MultilingualCO}. It is a collection of 33,807 news articles in three languages: English, Spanish and German. These articles are labeled by language and by story. Stories are multilingual, i.e. that they may contain articles from several languages. The training set contains 20,803 articles and the test set 13,004 articles. We further divide the training set in two: a train part to learn the $\beta$ weights of the linear combinations to aggregate similarities between articles (\ref{eq:equation1}) and a development part to set the hyper parameter $T_1$ and the resolution parameter of the Louvain algorithm (Section \ref{monolingualStories}) as well as threshold $T_2$ (Section \ref{fromMonoToMulti}). In order to set these parameters, we perform a grid search maximizing the average between standard and BCubed F1 scores. Table \ref{table:table1} presents descriptive statistics of the dataset.

\begin{table}[ht]
\begin{center}
\caption{Statistics for the train, development and test datasets.}
\label{table:table1}
\bigskip
\begin{tabular}{c|c|cccc}
\hline
\enspace\textbf{Partition}\enspace & \enspace\textbf{Language}\enspace & \textbf{\begin{tabular}[c]{@{}c@{}}\enspace Nb of \enspace \\ \enspace documents\enspace\end{tabular}} & \textbf{\begin{tabular}[c]{@{}c@{}}\enspace Avg nb of \enspace \\ \enspace words (std)\enspace\end{tabular}} & \textbf{\begin{tabular}[c]{@{}c@{}}\enspace Nb of \enspace \\ \enspace of clusters \enspace\end{tabular}} & \textbf{\begin{tabular}[c]{@{}c@{}}\enspace Avg cluster \enspace\\ \enspace size (std) \enspace\end{tabular}} \\ \hline
\multirow{4}{*}{Train} & English & 5,804 & 453 (359) & 296 & 20 (25) \\
                       & Spanish & 2,278 & 357 (219) & 208 & 11 (7) \\
                       & German & 1,879 & 286 (212) & 188 & 10 (6) \\
                       & All & 9,961 & 399 (315) & 557 & 18 (20) \\ \hline
\multirow{4}{*}{Dev} & English & 6,429 & 438 (397) & 297 & 22 (37) \\
                       & Spanish & 2,249 & 375 (264) & 208 & 11 (11) \\
                       & German & 2,164 & 295 (227) & 189 & 11 (7) \\
                       & All & 10,842 & 397 (348) & 557 & 19 (36) \\ \hline
\multirow{4}{*}{Test}  & English & 8,726 & 546 (518) & 222 & 39 (88) \\
                       & Spanish & 2,177 & 412 (358) & 149 & 15 (21) \\
                       & German & 2,101 & 458 (496) & 118 & 18 (45) \\
                       & All & 13,004 & 509 (494) & 381 & 34 (99) \\ \hline
\end{tabular}
\end{center}
\end{table}

\subsection{Results}
In order to assess the performances of our system, we report the standard as well as the BCubed\footnote{Unlike the classic version, the BCubed version of precision, recall and F1 score favors solutions that (i) make errors in clusters with already many errors (ii) make errors in a large clusters rather than in small ones.} \cite{Amig2008ACO} precision, recall and F1 score.
We evaluate our system for two tasks: monolingual and multilingual news clustering. %We assess the monolingual clustering part of our system for each language separately, using articles from the test set. From a multilingual perspective, we test our entire system, linking monolingual stories into multilingual ones.

\subsubsection{Monolingual results}
First, we can observe that for each language, our method produces a number of clusters closer to the reality. Then, on English documents, our method generates the best F1 and accuracy scores. However, when looking at BCubed metrics, we can see that our system ranks second after the method introduced in Staykovski \textit{et al.} \cite{Staykovski2019DenseVS}.
Nevertheless, for the two other languages which are German and Spanish, our method surpasses the system introduced in Miranda \textit{et al.} \cite{Miranda2018MultilingualCO}. Indeed, it displays better F1 and BCubed F1 scores, with improvements of 1.51 points for the F1 and 1.08 points for the BCubed F1 scores on German articles. Sometimes, even if our system is less precise than the one of Miranda \textit{et al.} \cite{Miranda2018MultilingualCO}, it displays a much higher recall, yielding better standard and BCubed F1 scores.
Notice that only 2\% of our articles are replayed using our "replaying" strategy. This insure that we are not performing a clustering from scratch each time we receive new articles.

\begin{table}[ht]
\begin{center}
\caption{Monolingual clustering results on the test dataset. F1 is F1 score, P stands for precision and R for recall. Best results are in bold.}
\label{table:table2}
\bigskip
\begin{tabular}{c|c|ccc|ccc|c}
\hline
\textbf{} & \textbf{Systems} & \multicolumn{3}{c|}{\textbf{Bcubed}} & \multicolumn{3}{c|}{\textbf{Standard}} & \textbf{Nb of} \\
 &  & \textbf{F1} & \textbf{P} & \textbf{R} & \textbf{F1} & \textbf{P} & \textbf{R} & \textbf{Clusters} \\ \hline
\multirow{4}{*}{English} & \textit{newsLens} & \enspace89.76\enspace & \enspace94.37\enspace & \enspace85.58\enspace & \enspace95.09\enspace & \enspace95.90\enspace & \enspace94.30\enspace & \enspace873\enspace \\
 & Miranda \textit{et al.} & 92.36 & 94.57 & 90.25 & 94.03 & 98.14 & 90.25 & 326 \\
 & Staykovski \textit{et al.}  & \textbf{94.41} & 95.16 & 93.66 & 98.11 & 97.60 & 98.63 & 484 \\
 & Ours & 93.86 & 94.19 & 93.55 & \textbf{98.31} & 98.21 & 98.42 & 298 \\ \hline
 
 \multirow{4}{*}{Spanish} & \textit{newsLens} & - & - & - & - & - & - & - \\
 & Miranda \textit{et al.} & 91.61 & 96.44 & 87.25 & 96.83 & 97.01 & 96.65 & 281 \\
 & Staykovski \textit{et al.} & - & - & - & - & - & - & - \\
 & Ours & \textbf{91.79} & 93.56 & 90.08 & \textbf{97.68} & 98.02 & 97.34 & 267\\
 \hline
 
\multirow{4}{*}{German} & \textit{newsLens} & - & - & - & - & - & - & - \\
 & Miranda \textit{et al.} & 93.64 & 98.92 & 88.90 & 97.19 & 99.86 & 94.67 & 229 \\
 & Staykovski \textit{et al.} & - & - & - & - & - & - & - \\
 & Ours & \textbf{94.72} & 95.13 & 94.31 & \textbf{98.70} & 99.16 & 98.24 & 205 \\ \hline
\end{tabular}
\end{center}
\end{table}

\subsubsection{Crosslingual results}
In the multilingual clustering setting, we compare our system to the one of Miranda \textit{et al.} \cite{Miranda2018MultilingualCO}, which is the only system handling multilingual news articles. We can see that our system displays better F1 (+2.5 points), precision and recall scores. This result shows that our improved system is able to both better organize monolingual stories, and link these stories over languages making use of our fine tuned DistilBERT embedding model.

\begin{table}[ht]
\begin{center}
\caption{Crosslingual clustering results on the test dataset. Best results are in bold.}
\label{table:table2}
\bigskip
\centering
\begin{tabular}{c|ccc|ccc|c}
\hline
\textbf{Systems} & \multicolumn{3}{c|}{\textbf{BCubed}} & \multicolumn{3}{c|}{\textbf{Standard}} & \textbf{Nb of} \\
\textbf{} & \textbf{F1} & \textbf{P} & \textbf{R} & \textbf{F1} & \textbf{P} & \textbf{R} & \textbf{Clusters} \\ \hline
Miranda \textit{et al.} & - & - & - & 84.0 & 83.0 & 85.0 & - \\
Ours & \enspace82.06\enspace & \enspace80.25\enspace & \enspace83.97\enspace & \enspace\textbf{86.49}\enspace & \enspace85.11\enspace & \enspace87.92\enspace & 606 \\ \hline
\end{tabular}
\end{center}
\end{table}
\section{Conclusion and Future Work}
We described a new method to cluster multilingual news articles into stories. We processe articles per batch as in \textit{newsLens}, and naturally link found topics along time by maintaining centroids for monolingual clusters. More precisely, we introduced a new "replaying" strategy to link monolingual topics into stories, and  then create crosslingual stories by embedding articles thanks to SBERT \cite{Reimers2019SentenceBERTSE}. Our system gives both monolingual and crosslingual state-of-the-art results on the English, Spanish and German dataset introduced by Miranda \textit{et al.} \cite{Miranda2018MultilingualCO}.
%More broadly, we use this news clustering system internally in order to automatically process financial news articles and track important events.

In future work, we plan to challenge the TF-IDF based representation of monolingual articles using fine tuned SBERT \cite{Reimers2019SentenceBERTSE} embeddings. Moreover, it would be interesting to assess computational efficiency of different systems by testing them on bigger news dataset.

\section*{Acknowledgements}

The authors would also like to thank Mr. Clément Rebuffel, Mr. Pirashanth Ratnamogan, and Mr. Bruce Delattre from BNP Paribas for their valuable comments and suggestions.

\bibliographystyle{alpha} 
\bibliography{biblio}

\newcommand{\etalchar}[1]{$^{#1}$}
\begin{thebibliography}{SBCMN19}

\bibitem[AGAV08]{Amig2008ACO}
Enrique Amig{\'o}, Julio Gonzalo, Javier Artiles, and M.~Felisa Verdejo.
\newblock A comparison of extrinsic clustering evaluation metrics based on
  formal constraints.
\newblock volume~12, pages 461--486, 2008.

\bibitem[BGLL08]{Blondel2008FastUO}
Vincent~D. Blondel, Jean-Loup Guillaume, Renaud Lambiotte, and Etienne
  Lefebvre.
\newblock Fast unfolding of communities in large networks.
\newblock 2008.

\bibitem[Kuh55]{Kuhn1955Hungarian}
Harold~W. Kuhn.
\newblock {The Hungarian Method for the Assignment Problem}.
\newblock volume~2, pages 83--97, March 1955.

\bibitem[LH17]{Laban2017newsLensBA}
Philippe Laban and Marti~A. Hearst.
\newblock newslens: building and visualizing long-ranging news stories.
\newblock In {\em NEWS@ACL}, 2017.

\bibitem[LM14]{Le2014DistributedRO}
Quoc~V. Le and Tomas Mikolov.
\newblock Distributed representations of sentences and documents.
\newblock volume 1405.4053, 2014.

\bibitem[MZCB18]{Miranda2018MultilingualCO}
Sebasti{\~a}o Miranda, Arturs Znotins, Shay~B. Cohen, and Guntis Barzdins.
\newblock Multilingual clustering of streaming news.
\newblock In {\em EMNLP}, 2018.

\bibitem[RG19]{Reimers2019SentenceBERTSE}
Nils Reimers and Iryna Gurevych.
\newblock Sentence-bert: Sentence embeddings using siamese bert-networks.
\newblock In {\em EMNLP/IJCNLP}, 2019.

\bibitem[RML{\etalchar{+}}15]{Rupnik2015NewsAL}
Jan Rupnik, Andrej Muhic, Gregor Leban, Blaz Fortuna, and Marko Grobelnik.
\newblock News across languages - cross-lingual document similarity and event
  tracking.
\newblock In {\em IJCAI}, 2015.

\bibitem[SBCMN19]{Staykovski2019DenseVS}
Todor Staykovski, Alberto Barr{\'o}n-Cede{\~n}o, Giovanni Da~San Martino, and
  Preslav Nakov.
\newblock Dense vs. sparse representations for news stream clustering.
\newblock In {\em Text2Story@ECIR}, 2019.

\bibitem[SDCW19]{Sanh2019DistilBERTAD}
Victor Sanh, Lysandre Debut, Julien Chaumond, and Thomas Wolf.
\newblock Distilbert, a distilled version of bert: smaller, faster, cheaper and
  lighter.
\newblock {\em ArXiv}, abs/1910.01108, 2019.

\end{thebibliography}
%inline the .bbl file directly for mailing to authors.

%\begin{thebibliography}{Com79}

%\bibitem[Com79]{Comer-btree}
%D.~Comer.
%\newblock The ubiquitous b-tree.
%\newblock {\em Computing Surveys}, 11(2):121--137, June 1979.

%\bibitem[Knu73]{Knuth-vol3}
%D.~E. Knuth.
%\newblock {\em The Art of Computer Programming -- Volume 3 / Sorting and
%  Searching}.
%\newblock Addison-Wesley, 1973.

%\end{thebibliography}

\end{document}